\documentclass{article}
\usepackage{spconf,amsmath,graphicx}
\usepackage{amssymb}
\usepackage[noend]{algpseudocode}
\usepackage{algorithm}
\usepackage{graphicx}
\usepackage{textcomp}
\usepackage{comment}
\usepackage{xcolor}
\usepackage{placeins}
\usepackage{amsmath,amsfonts,amsthm,bm}
\usepackage{multirow}
\usepackage{pdflscape}
\usepackage{tabularx}
\usepackage{lipsum}
\usepackage{caption} 
\usepackage{subcaption}
\usepackage{csquotes}
\usepackage{comment}
\usepackage{hyperref}
\usepackage{bbm}

\usepackage{algpseudocode,algorithm,booktabs}

\algnewcommand{\Inputs}[1]{%
  \State \textbf{Inputs:}
  \parbox[t]{.8\linewidth}{\raggedright #1}
}
\algnewcommand{\Initialize}[1]{%
  \State \textbf{Initialize:}
  \parbox[t]{.8\linewidth}{\raggedright #1}
}
\algnewcommand{\LeftComment}[1]{\Statex \(\triangleright\) #1}

\newcommand{\corr}{\textsc{correction}}
\newcommand{\coreg}{\textsc{coregistration}}
\def\BibTeX{{\rm B\kern-.05em{\sc i\kern-.025em b}\kern-.08em
    T\kern-.1
    7em\lower.7ex\hbox{E}\kern-.125emX}}
    \newcommand{\bt}[1]{\mbox{$\bf #1$}}

\title{GPU-accelerated SIFT-aided source identification of stabilized videos}

\begin{document}
\ninept
\maketitle
\begin{abstract}

Video stabilization is an in-camera processing commonly applied by modern acquisition devices. While significantly improving the visual quality of the resulting videos, it has been shown that such operation typically hinders the forensic analysis of video signals. In fact, the correct identification of the acquisition source usually based on Photo Response non-Uniformity (PRNU) is subject to the estimation of the transformation applied to each frame in the stabilization phase. A number of techniques have been proposed for dealing with this problem, which however typically suffer from a high computational burden due to the grid search in the space of inversion parameters.
Our work attempts to alleviate these shortcomings by exploiting the parallelization capabilities of Graphics Processing Units (GPUs), typically used for deep learning applications, in the framework of stabilised frames inversion. Moreover, we propose to exploit SIFT features {to estimate the camera momentum and} 
identify less stabilized temporal segments, thus enabling a more accurate identification analysis, and to efficiently initialize the frame-wise parameter search of consecutive frames.
Experiments on a consolidated benchmark dataset confirm the effectiveness of the proposed approach in reducing the required computational time and improving the source identification accuracy. {The code is available at \url{https://github.com/AMontiB/GPU-PRNU-SIFT}}.
\end{abstract}
\begin{keywords}
Video Source Identification, GPU, SIFT, PRNU, Video Stabilization
\end{keywords}
\section{Introduction}
\label{sec:intro}
Digital devices, social media platforms and multimedia data have an increasingly relevant role in our daily life. In order to avoid content improper usage and spreading, it is necessary to develop techniques to prove their origin \cite{lukas2006digital,cozzolino2018noiseprint} and identify their source. 
By source identification we refer to the procedure, mostly required in investigations and courtroom cases, in which the device (i.e., the camera or smartphone) that acquired an image or a video under investigation is identified. With this regard, the research literature is particularly rich when it comes to techniques dealing with images \cite{lukas2006digital, dirik2007source, fridrich2009digital}, among which the most reliable ones use the so-called Photo Response non-Uniformity (PRNU) \cite{lukas2006digital}, a scant and unique residual introduced by the camera sensor every time an image or a video is taken. 
In fact, by extracting the PRNU from two different images and comparing them in terms of Peak of Correlation Energy (PCE), it can be verified whether they were taken with the same device or not.

However, the PRNU is highly sensitive to spatial transformations (such as radial corrections \cite{Goljian2012Sensor}, digital zoom \cite{Goljan2008Digital}, HDR correction \cite{Darvish2019Camera}), as they cause a misalignment of such noise patterns. In this case, the inverse spatial transformation must be applied in order to restore the original pattern and reliably compare it with reference ones; this requires the estimation of the transformation parameters applied in the first place.
Such issues are even more impactful when applied to video source identification, due to stronger compression and more complex spatial transformations such as the Electronic Image Stabilization (EIS) \cite{morimoto1998evaluation}, applied by modern devices to improve video quality. 

To invert EIS transformations and restore the reliability of the PRNU, many works propose to use a combination of grid searches, predicting methods and parallel CPU processing \cite{mandelli2019facing, mandelli2020modified, iuliani2019hybrid}. However, a common trait of such approaches is the rather high computational burden they entail. 
In \cite{iuliani2019hybrid},  Iuliani et. al. check every possible combination of scaling, rotation and shift parameters by means of a grid search on each frame; to reduce the computational cost, some of the parameters are estimated offline and used as a-priori information. In \cite{mandelli2019facing}, Mandelli et. al. propose a faster algorithm for the inversion of the EIS, which however implies significant hardware requirements to be computationally efficient.
Finally, in \cite{mandelli2020modified} Mandelli et. al. propose a method based on a modified version of the Fourier-Mellin transform for efficient estimation of the rotation parameter and the inversion of the EIS. The algorithm obtains promising results in terms of accuracy and computational cost but (just like \cite{mandelli2019facing, iuliani2019hybrid}) it exploits only the information coming from the video intra-frames (I), while fully discarding P and B frames.
Furthermore, although the EIS transformations are typically modelled through 8 parameters, all the cited methods (\cite{mandelli2019facing, mandelli2020modified, iuliani2019hybrid}) target the estimation of only three of them in order to avoid combinatorial explosion, thus decreasing the inversion accuracy. 

In performing source identification, an alternative approach to the conventional PRNU extraction is the computation of a proper residual by means of deep neural networks, whose weights are learned through a training procedure. It is the case of Noiseprint \cite{cozzolino2018noiseprint}, which has been successfully applied for digital images and recently extended to videos \cite{cozzolino2019extracting}, although not dealing with video stabilization issues.


Given the current limitations of existing approaches, in this paper, we propose an innovative solution for source identification of stabilized videos using the PRNU. Our algorithm inverts the EIS by pre-selecting the less stabilized frames through a blind camera momentum estimator before estimating the inversion parameters via grid search.
In this phase, we leverage the higher computational and parallelization capabilities of the GPU architectures, which particularly fit our needs as they are optimized for similar point-wise operations arising in computer graphics applications and act here as computing accelerators. Our pipeline includes the use of SIFT features{- already used in forensics \cite{bellavia2019prnu, bellavia2021experiencing} but never for source identification -} in order to exploit the temporal correlation between neighbour frames and efficiently initialize their search parameters.

The theoretical background behind our work is described
in Section \ref{sec:math_model} and the proposed method is detailed in Section \ref{sec:proposed_solution}. Experimental results and comparison with the literature are discussed in Section \ref{sec:experiment}, while future directions and conclusions are drawn in Section \ref{sec:conclusions}.

\section{Theoretical Background}
\label{sec:math_model}
\subsection{\textbf{Notation}}
In this paper, $M\times N$ matrices will be denoted with uppercase boldface letters $\bt X$, their $(i^{th}, j^{th})$ elements as $X_{i,j}$, and their mean value over all elements as $\overline{\bt X}$. Similarly, $M$-dimensional vectors are denoted as lowercase boldface letters $\bt x$ with mean value $\overline{\bt x}$.
We indicate with $\odot$ the dot product of vectors and with $\mathbbm{1}$ the identity matrix. 

For a video under analysis, the $u$-th I-frame appearing in the video stream is denoted as $\bt{I}_u$. 
The normalized cross-correlation (NCC) between two matrices $\bt X$ and $\bt Y$ (of the same size $M \times N$) is defined as follows \cite{Chen2008Determining}:
\begin{equation}
\label{eq:NCC}
 \rho(\bt X, \bt Y)= \frac{(\bt X- \overline{\bt X})\odot(\bt Y- \overline{\bt Y})}{||\bt X- \overline{\bt X}|| \cdot ||\bt Y- \overline{\bt Y}||}
 \end{equation}
 If $\bt X$ and $\bt Y$ have different sizes 
 we apply zero-padding to the smaller one \cite{Goljan2008Digital}.

\subsection{Photo Response non-Uniformity}
The PRNU (Photo-Response Non-Uniformity) is a scant residual introduced by the acquiring sensor when the light hits its components, caused by small variations of the output signal from pixel to pixel. Such aberration is unique, related to the materials and manufacturing process of the sensor and more detectable in brighter flat areas \cite{Goljian2014Estimation, Goljan2008Digital}.
The PRNU is a very weak signal modelled as multiplicative noise \cite{Goljan2009Large} and has to be separated from other noise components when extracting it from the image or frame under analysis. 
%
%
Typically, the {\it noise residual} containing the PRNU is estimated from a single image as $W(\bt I) \doteq \bt I-F(\bt I)$
where $F(\cdot)$ is the denoiser \cite{cortiana2011performance, amerini2009analysis}, which in our case is the Mihcak's wavelet-based denoiser \cite{Mihcak1999Denoiser}. 
Instead, the {\it reference fingerprint} for a given device is computed starting from $L$ images $\bt I^{(l)}$, $1=1, \cdots, L$ taken from the device
as follows \cite{Goljan2009Large}:
\begin{equation}
	{\hat {\bt K}}=\left(\sum_{l=1}^L \bt I^{(l)} \cdot W(\bt I^{(l)}) \right)\cdot \left(\sum_{l=1}^L \bt I^{(l)} \cdot \bt I^{(l)} \right)^{-1}
	\label{eq:camera_fingeprint}
\end{equation}
In Eq. \eqref{eq:camera_fingeprint}, all the operations are pixel-wise.
\par Given a test image $\bt I$, the source identification is accomplished by solving a binary hypothesis test consisting in verifying whether $\bt I$ contains the same PRNU as the camera fingerprint $\hat {\bt K}$. We will denote the null hypothesis of this test (i.e., $\bt I$ does not contain $\hat{\bt K}$) by $H_0$ and the alternative one (i.e., $\bt I$ contains $\hat{\bt K}$) by $H_1$. We verify these hypotheses using as test statistics the Peak-to-Correlation Energy ratio (PCE) defined in Eq. \eqref{eq:pce}, which consists in computing the peak cross-correlation between the test image residual $W(\bt I)$ and the camera fingerprint $\hat{\bt K}$ and normalizing it by an estimate of the correlation noise under $H_0$ \cite{Kang12}: 
\begin{equation}
    \text{PCE}(\hat{\bt K},\bt I)=\frac{\text{sgn}(\rho(\hat{\bt K},W(\bt I)_{\boldsymbol \delta_{\text{peak}}})) \cdot \rho^2(\hat{\bt K}, W(\bt I)_{\boldsymbol \delta_{\text{peak}}})}{\frac{1}{MN-|{\mathcal{D}}|}\sum_{\boldsymbol{\delta} \in {\mathcal I} \backslash {\mathcal{D}}} \rho^2(\hat{\bt K}, W(\bt I)_{i+\delta_1, j+\delta_2})}
   \label{eq:pce}
\end{equation}
\noindent
where $\boldsymbol{\delta}= (\delta_1, \delta_2)$ are all the possible shifts that occur between $W(\bt I)$ of size $M'\times N'$ and $\hat{\bt K}$ of size $M \times N$ such that $0 \leq \delta_1 \leq M-M'\quad \text{and}\quad 0\leq \delta_2\leq N-N'$ and $\boldsymbol \delta_{\text{peak}}$ are the coordinates of the peak of correlation according to \cite{Goljan2008Camera}. 
$\mathcal I$ defines all possible image pixels coordinates, while
$\mathcal D$ is a cyclic exclusion neighbourhood around the peak of correlation value of size $11 \times 11$ pixels to avoid contamination of cross-correlation peaks from $H_1$ when estimating the cross-correlation noise under $H_0$ \cite{Goljan2008Digital}.

\subsection{\textbf{Electronic Image Stabilization }}
\label{ssec:eis}
Electronic Image Stabilization (EIS) \cite{morimoto1998evaluation} is a post-processing technique used in modern devices and cameras to stabilize video sequences by compensating for temporal camera motion. To this end, a spatial transformation is applied to the acquired frames, which entails a parametric coordinate mapping followed by interpolation. We can model as follows the inverse coordinate mapping between the stabilized and the original frame pixels: 
\begin{equation}
    \label{eq:model}
    \begin{pmatrix}
        w\\
        h\\
        1
    \end{pmatrix}
    = \bt  T_{\bt t} \cdot \begin{pmatrix}
        w' \\
        h' \\
        1
    \end{pmatrix}
    = 
    \begin{bmatrix} 
    t_{1,1} & t_{1,2} & t_{1,3}\\
    t_{2,1} & t_{2,2} & t_{2,3} \\
    t_{3,1} & t_{3,2} & 1
    \end{bmatrix}
    \cdot
    \begin{pmatrix}
        w' \\
        h' \\
        1
    \end{pmatrix}
\end{equation}
where $(w',h')$ are the coordinates of the stabilized pixels and $(w,h)$ the original ones, while $\bt t = [t_{1,1}, t_{1,2}, ..., t_{3,2}]$ is a 8-dimensional vector of varying parameters. In particular, $t_{1,1}$ and $t_{2,2}$ are related to horizontal/vertical scaling, $t_{1,2}$ and $t_{2,1}$ to rotation, $t_{1,3}$ and $t_{2,3}$ to translation, and finally $t_{3,1}$ and $t_{3,2}$ are the projective parameters. Such model encompasses different types of EIS systems \cite{morimoto1998evaluation}, operating on 3-, 5- or 6-axes, depending on the nature of the transformations. With a slight abuse of notation, if $\bt Y$ is a generic video frame, we will write $\bt  T_{\bt t}(\bt Y)$ to indicate the version of $\bt Y$ whose pixels underwent the grid transformation with parameters $\bt t$ followed by interpolation.

\subsection{Keypoint matching and homography estimation}
\label{ssec:kp}

\begin{figure}[h!]
    \centering
    \includegraphics[width=0.45\textwidth]{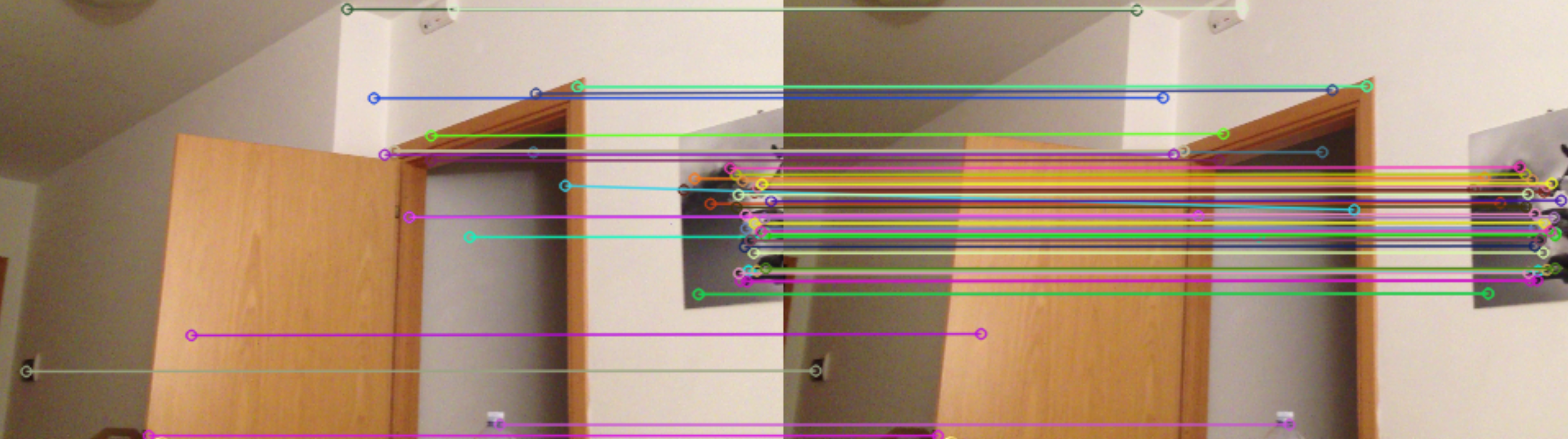}
    \caption{\footnotesize Example of two frames where SIFT keypoints are detected (colored circles) and matched through the DEGENSAC algorithm (colored lines.)}
    \label{fig:kp}
\end{figure}

The homography relation between two frames can be estimated through the detection and the interframe matching of keypoints. In particular, we can associate to a pair of two generic frames $\bt X$ and $\bt Y$:
\begin{itemize}
    \item $\mathcal{S}$, a set containing pairs  $(\bt s_{\bt X}, \bt s_{\bt Y})$ of 2-D SIFT keypoints \cite{lowe2004distinctive} that have been detected in $\bt X$ and $\bt Y$, respectively, and result as matching from the application of the DEGENSAC algorithm \cite{chum2005two}. An example of this detection and matching  process is reported in Figure \ref{fig:kp}. We employ the \textit{Open-CV} libraries for this purpose;
    
    \item $\bt H_\mathcal{S}$, the estimated homography matrix between $\bt X$ and $\bt Y$, which is provided as a by-product by the DEGENSAC algorithm starting from the keypoints in $\mathcal{S}$. $\bt H_{\mathcal{S}}$ has a similar model as in Eq. \eqref{eq:model} and, by using the same notation convention, we expect $\bt X \approx \bt H_{\mathcal{S}} (\bt Y)$. 
\end{itemize}

In addition, we also define $\tilde{\mathcal{S}}$, a sanitized set of matching keypoints where only those yielding an interframe Euclidean distance $|| \bt s_{\bt X} - \bt s_{\bt Y} ||_2$ below an empirical threshold are retained.

\section{Proposed Solution}
\label{sec:proposed_solution}

We consider the hybrid scenario where the reference fingerprint of a device is estimated starting from flat images acquired by the same device, as in \eqref{eq:camera_fingeprint}.  

As highlighted in previous approaches, in order for it to be used for testing video frames, the image-based fingerprint needs to be properly down-scaled and cropped due to size mismatch between image and video acquisition; we perform this operation similarly to what is done in \cite{mandelli2019facing}, so to obtain a fingerprint $\hat{\bt K}$ to be used for testing video frames.


For a generic frame $\bt I$, the core of the identification analysis consists in searching for the parameters $\bt t$ that maximize the PCE value as in Eq. \eqref{eq:pce} between $\hat{\bt K}$ and the residual extracted from $\bt T_{\bt t}(\bt I)$, the latter being as close as possible to the originally acquired frame before the stabilization. 

In order to improve the efficiency and the accuracy of this process, we propose a two-phase methodology encompassing a preselection of lightly stabilized frames (described in Section \ref{subsec:bestframe}) and a frame-wise inversion analysis boosted by a SIFT-based homography estimation (described in Section \ref{subsec:eis_inv}). 
Moreover, we developed a Tensorflow implementation of the overall procedure, building on the Tensorflow add-on libraries for the frame-wise inversion operations, and the PCE and the cross-correlation formulas defined in Eqs. \eqref{eq:pce} and \eqref{eq:NCC}, respectively.

\subsection{Selection of low-stabilization frames}
\label{subsec:bestframe}

In this first phase, pairs of consecutive I-frames are analyzed, with the goal of locating the Group of Pictures (GOPs) where the weaker stabilization has supposedly been applied. Inspired by \cite{battiato2007sift}, we achieve this by computing a {\it camera momentum}, which expresses the global amount of motion between frames. We interpret this measure as a proxy for the strength of the stabilization operation applied to the originally acquired frame: our intuition is that less stabilized frames yield more reliable frame inversion and PRNU matching processes. 

Given two consecutive I-frames $\bt I_{u}$ and $\bt I_{u+1}$, the set $\tilde{\mathcal{S}_u}$ containing sanitized matching keypoint pairs $(\bt s_{u},\bt s_{u+1})$ is obtained as described in Section \ref{ssec:kp}. The camera momentum between  $\bt I_{u}$ and $\bt I_{u+1}$ is then defined as
\begin{equation}
\label{eq:momentum}
    \overline{\mathbf{\Delta}}_u \doteq \sum\limits_{\tiny (\bt s_{u}, \bt s_{u+1}) \in \tilde{\mathcal{S}_u} } ||\bt s_{u} - \bt s_{u+1}||_2 \, \cdot \, |\tilde{\mathcal{S}_u}|^{-1},
\end{equation}
that is, the average interframe displacement between matching keypoint pairs in $\tilde{\mathcal{S}_u}$.

By iterating this operation along the video duration, we can identify the index $A$ such that
\begin{equation}
A =  \arg \min_{u} \overline{\mathbf{\Delta}}_u
\label{eq:anchor_frames}
\end{equation}
The corresponding I-frame $\bt I_A$ is defined as the {\it anchor} and identifies the starting point of the successive frame-wise inversion analysis, which will be limited to the frames
\begin{equation} \label{eq:sel}
    \bt I_A, \bt P_{1}, \bt P_{2}, \ldots ,\bt P_{V_A}
\end{equation}
where $\bt P_v$, $v=1, \ldots, V_A$ are predicted frames (P or B type) except for $\bt P_{V_A} \equiv \bt I_{A+1}$, and $V_A$ is the GOP size. When the set $\tilde{\mathcal{S}_u}$ is empty, as for flat videos, and $\overline{\mathbf{\Delta}}_u$ cannot be estimated, the index $A$ corresponds to the first I-frame of the video.

\subsection{SIFT-aided EIS Inversion}
\label{subsec:eis_inv}

In this second phase, we aim at filling a vector $\boldsymbol{\gamma} = [\gamma_A, \gamma_{1}, \ldots, \gamma_{V_A} ]$ containing the maximum 
PCE value with the reference fingerprint $\hat{\bt K}$ measured at each of the selected frame indices in Eq. \eqref{eq:sel} under a number of tested inverse transformations of the frame. 

\begin{algorithm}[b!]
\scriptsize
\caption{SIFT-aided EIS inversion}\label{alg:cap}
\begin{algorithmic}[1]
\Inputs{anchor index $A$, \\ frames $\bt I_{A}$ , $\bt P_{A+1}, \ldots ,\bt P_{A+V_A}$\\ reference fingerprint $\hat{\bt K}$}

\vspace{0.2cm}

\Initialize{ vector $\boldsymbol{\gamma} $, temporary frame $\bt U$}

\vspace{0.2cm}

\LeftComment{{\it Correction of the anchor}}
\State $(\bt I_{A}^{(c)}, \gamma_A) \gets $ {\corr}$(\hat{\bt K}, \bt I_A)$
\State $\bt U \gets \bt I_{A}^{(c)} $

\vspace{0.2cm}

\LeftComment{{\it Co-registration and correction of the successive frames}}
\For{$v=1, \ldots, V_A$}


\State $(\bt H_{v},\bt P_{v}^{(r)}) \gets $ {\coreg}$(\bt U,\bt P_{v})$


\If{PCE$(\hat{\bt K}, \bt P_{v}^{(r)}) > $ PCE$(\hat{\bt K},\bt P_{v})$}
\State $(\bt P_{v}^{(c)},\gamma_v)= $ {\corr$(\hat{\bt K}, \bt P_{v}, \bt H_{v})$}
\Else
\State $(\bt P_{v}^{(c)},\gamma_v)= $ {\corr}$(\hat{\bt K},\bt P_{v})$
\EndIf

\State $\bt U \gets \bt P_{v}^{(c)}$

\EndFor

\vspace{0.2cm}

\State {\bf return} vector $\boldsymbol{\gamma}$

\end{algorithmic}

\end{algorithm}

\begin{table*}[ht!]
\centering
\resizebox{17cm}{!}{%
\begin{tabular}{ccccccccccccccccccccccccc}
\cline{2-25}
\multicolumn{1}{c|}{\multirow{2}{*}{}}                                                  & \multicolumn{2}{c|}{D02}                                              & \multicolumn{2}{c|}{D05}                                              & \multicolumn{2}{c|}{D06}                                                 & \multicolumn{2}{c|}{D10}                                              & \multicolumn{2}{c|}{D14}                                              & \multicolumn{2}{c|}{D15}                                              & \multicolumn{2}{c|}{D18}                                              & \multicolumn{2}{c|}{D19}                                                 & \multicolumn{2}{c|}{D20}                                              & \multicolumn{2}{c|}{D25}                                              & \multicolumn{2}{c|}{D29}                                              & \multicolumn{2}{c|}{D34}                                              \\
\multicolumn{1}{c|}{}                                                                   & \multicolumn{1}{c|}{TPR}        & \multicolumn{1}{c|}{ETPS}           & \multicolumn{1}{c|}{TPR}        & \multicolumn{1}{c|}{ETPS}           & \multicolumn{1}{c|}{TPR}           & \multicolumn{1}{c|}{ETPS}           & \multicolumn{1}{c|}{TPR}        & \multicolumn{1}{c|}{ETPS}           & \multicolumn{1}{c|}{TPR}        & \multicolumn{1}{c|}{ETPS}           & \multicolumn{1}{c|}{TPR}        & \multicolumn{1}{c|}{ETPS}           & \multicolumn{1}{c|}{TPR}        & \multicolumn{1}{c|}{ETPS}           & \multicolumn{1}{c|}{TPR}           & \multicolumn{1}{c|}{ETPS}           & \multicolumn{1}{c|}{TPR}        & \multicolumn{1}{c|}{ETPS}           & \multicolumn{1}{c|}{TPR}        & \multicolumn{1}{c|}{ETPS}           & \multicolumn{1}{c|}{TPR}        & \multicolumn{1}{c|}{ETPS}           & \multicolumn{1}{c|}{TPR}        & \multicolumn{1}{c|}{ETPS}           \\ \hline
\multicolumn{1}{c|}{\begin{tabular}[c]{@{}c@{}}Ours\\ $\tau_{0.05}=19.5$\end{tabular}}  & \multicolumn{1}{c|}{\textbf{1}} & \multicolumn{1}{c|}{{12.27}} & \multicolumn{1}{c|}{\textbf{1}} & \multicolumn{1}{c|}{{11.72}} & \multicolumn{1}{c|}{0.72}          & \multicolumn{1}{c|}{{9.85}} & \multicolumn{1}{c|}{\textbf{1}} & \multicolumn{1}{c|}{{16.56}} & \multicolumn{1}{c|}{0.92}       & \multicolumn{1}{c|}{{19.77}} & \multicolumn{1}{c|}{\textbf{1}} & \multicolumn{1}{c|}{{23.29}} & \multicolumn{1}{c|}{\textbf{1}} & \multicolumn{1}{c|}{{13.34}} & \multicolumn{1}{c|}{\textbf{0.91}} & \multicolumn{1}{c|}{{11.30}} & \multicolumn{1}{c|}{\textbf{1}} & \multicolumn{1}{c|}{{11.69}} & \multicolumn{1}{c|}{0.64}      & \multicolumn{1}{c|}{{10.47}} & \multicolumn{1}{c|}{\textbf{1}} & \multicolumn{1}{c|}{{4.94}} & \multicolumn{1}{c|}{\textbf{1}} & \multicolumn{1}{c|}{{8.21}} \\ \hline
\multicolumn{1}{c|}{\begin{tabular}[c]{@{}c@{}}{M2019}\\ $\tau_{0.05}=36$\end{tabular}} & \multicolumn{1}{c|}{0.87}       & \multicolumn{1}{c|}{61.61}        & \multicolumn{1}{c|}{0.62}       & \multicolumn{1}{c|}{54.08}         & \multicolumn{1}{c|}{\textbf{0.88}} & \multicolumn{1}{c|}{52.33}         & \multicolumn{1}{c|}{0.87}       & \multicolumn{1}{c|}{51.47}           & \multicolumn{1}{c|}{0.87}       & \multicolumn{1}{c|}{51.46}         & \multicolumn{1}{c|}{0.63}       & \multicolumn{1}{c|}{60.65}         & \multicolumn{1}{c|}{0.5}        & \multicolumn{1}{c|}{47.21}         & \multicolumn{1}{c|}{0.75}          & \multicolumn{1}{c|}{38.27}         & \multicolumn{1}{c|}{0.88}       & \multicolumn{1}{c|}{37.97}           & \multicolumn{1}{c|}{\textbf{1}} & \multicolumn{1}{c|}{37.88}         & \multicolumn{1}{c|}{0.63}       & \multicolumn{1}{c|}{53.21}         & \multicolumn{1}{c|}{0.57}       & \multicolumn{1}{c|}{44.88}         \\ \hline
\multicolumn{1}{c|}{\begin{tabular}[c]{@{}c@{}}{MFM}\\ $\tau_{0.05}=34$\end{tabular}} & \multicolumn{1}{c|}{0.89}       & \multicolumn{1}{c|}{110.13}         & \multicolumn{1}{c|}{0.89}       & \multicolumn{1}{c|}{107.33}           & \multicolumn{1}{c|}{0.78}          & \multicolumn{1}{c|}{95.05}         & \multicolumn{1}{c|}{0.89}       & \multicolumn{1}{c|}{78.42}         & \multicolumn{1}{c|}{\textbf{1}} & \multicolumn{1}{c|}{72.29}         & \multicolumn{1}{c|}{0.78}       & \multicolumn{1}{c|}{66.32}         & \multicolumn{1}{c|}{0.89}       & \multicolumn{1}{c|}{76.50}        & \multicolumn{1}{c|}{0.89}          & \multicolumn{1}{c|}{57.54}         & \multicolumn{1}{c|}{\textbf{1}} & \multicolumn{1}{c|}{51.97}         & \multicolumn{1}{c|}{\textbf{1}} & \multicolumn{1}{c|}{49.50}        & \multicolumn{1}{c|}{0.67}       & \multicolumn{1}{c|}{37.58}           & \multicolumn{1}{c|}{0.55}       & \multicolumn{1}{c|}{39.88}         \\ \hline
\multicolumn{1}{c|}{\begin{tabular}[c]{@{}c@{}}{Ours CPU}\\ {$\tau_{0.05}=19.5$}\end{tabular}}  & \multicolumn{1}{c|}{{\textbf{1}}} & \multicolumn{1}{c|}{{615.19}} & \multicolumn{1}{c|}{{\textbf{1}}} & \multicolumn{1}{c|}{{492.42}} & \multicolumn{1}{c|}{{{0.72}}}          & \multicolumn{1}{c|}{{538.82}} & \multicolumn{1}{c|}{{\textbf{1}}} & \multicolumn{1}{c|}{{519.15}} & \multicolumn{1}{c|}{{\textbf{0.92}}}       & \multicolumn{1}{c|}{{554.55}} & \multicolumn{1}{c|}{{\textbf{1}}} & \multicolumn{1}{c|}{{611.62}} & \multicolumn{1}{c|}{{\textbf{1}}} & \multicolumn{1}{c|}{{566.81}} & \multicolumn{1}{c|}{{\textbf{0.91}}} & \multicolumn{1}{c|}{{558.43}} & \multicolumn{1}{c|}{{\textbf{1}}} & \multicolumn{1}{c|}{{563.09}} & \multicolumn{1}{c|}{{0.636}}      & \multicolumn{1}{c|}{{469.04}} & \multicolumn{1}{c|}{{\textbf{1}}} & \multicolumn{1}{c|}{{408.81}} & \multicolumn{1}{c|}{{\textbf{1}}} & \multicolumn{1}{c|}{{523.74}} \\ \hline

\end{tabular}
}
\vspace{-2mm}
\caption{\footnotesize Results obtained by the proposed method ``Ours", M2019 \cite{mandelli2019facing} and  MFM \cite{mandelli2020modified} in terms of TPR and ETPS (Elaboration Time Per Second of video) for a FPR=$0.05$ on the different devices. In \textbf{boldface} are highlighted the best results, the time is expressed in seconds.}
\label{tab:TRP_time}
\end{table*}

For this purpose, we define the following operators: 

\begin{itemize}
    
    \item {\corr}$(\hat{\bt K},\bt I,\bt T_{\text{init}})$: 
    given a reference PRNU fingerprint $\hat{\bt K}$ and a frame $\bt I$, this operator applies a breadth-first search \cite{bundy1984breadth} over the parameters in $\bt t$
    with the goal of maximizing $\text{PCE}(\hat{\bt K},\bt T_{\bt t}(\bt I))$. 
    The search starts from the parameters contained in the optional input matrix $\bt T_{\text{init}}$, if given.

    As in \cite{mandelli2019facing}, simplifying assumptions are made, in particular on the scaling parameters ($t_{1,1}= t_{2,2} \doteq \lambda$  ) and on the rotation parameters ($t_{1,2}= -t_{2,1}\doteq \theta$). Moreover, $t_{3,1}, t_{3,2}$ are set to $0$, and $t_{1,3}, t_{2,3}$ are estimated once and kept fixed. 
    
    At each $n$-th iteration, the pair $(\lambda^{(n)},\theta^{(n)})$ is determined through exhaustive search over $\bt \Lambda^{(n)} \times \bt \Theta^{(n)}$ as the one yielding the transformation of $\bt I$ with the highest PCE value. 
    
    The sets $\bt \Lambda^{(n)}$ and $\bt \Theta^{(n)}$ are finite and progressively narrower neighborhoods of the previous estimates. In particular: 
\begin{equation}
\label{eq:sets}
\begin{gathered}
\bt \Lambda^{(n)} = \{ \lambda^{(n-1)} + \alpha \cdot 0.01^{-n}\}_{\alpha \in {\mathbb Z} \cap [-5, 5]}\\
\bt \Theta^{(n)} = \{ \theta^{(n-1)} + \alpha \cdot 0.1^{-n}\}_{\alpha \in {\mathbb Z} \cap [-5, 5]}
\end{gathered}
\end{equation}
    {where $\alpha$ sets the size of $\bt \Lambda^{(n)}$ and $\bt \Theta^{(n)}$.} At the first iteration ($n=1$), if the matrix $\bt T_{\text{init}}$ is given as input, then a corresponding vector $\bt t_{\text{init}}$ is derived, from which $\lambda^{(0)}$ and $\theta^{(0)}$ are extracted accordingly. Otherwise, they are initialized to $1$ and $0$, respectively. Moreover, at the first iteration $\theta^{(1)}$ is searched on a denser grid ($\{-5, -4.9, \ldots, 4.9, 5 \}$), to improve accuracy.
    
    If at the $n$-th iteration, $(\lambda^{(n)},\theta^{(n)})\equiv(\lambda^{(n-1)},\theta^{(n-1)}) $, the process stops. Also, in our experiments we fixed the maximum number of iterations at $3$. $\bt t_{\text{max}}$ is the parameter vector for which the maximum PCE value $\gamma$ is observed.
    $\gamma$ is returned as output together with the corrected frame $\bt I^{(c)}$ 
    obtained by transforming $\bt I$ with $\bt t_{\text{max}}$.
    
    \item {\coreg}$(\bt I,\bt P)$: this operator returns as output the estimated homography matrix $\bt H$ between $\bt I$ and $\bt P$ computed as in Section \ref{ssec:kp}, and the resulting co-registered frame $\bt P^{(r)}\doteq \bt H (\bt P) \approx \bt I$. 
    
\end{itemize}

Those operators are combined in our proposed method as formalized in Algorithm \ref{alg:cap}. 
Essentially, the iterative correction procedure is applied to 
each frame. The overall idea is to first correct the anchor $\bt 
I_A$ so to obtain $\bt I_{A}^{(c)}$; then, the successive frames 
$\bt P_v$ are co-registered with respect to the previous one
prior to  correction, obtaining for each of them a registered 
version $\bt P_v^{(r)}$. If $\bt P_v^{(r)}$ yields a higher PCE 
value than $\bt P_v$, the correction is initialized by taking 
into account the homography estimation.

\par The final decision is taken by thresholding with a value $\tau$ the mean of the vector $\boldsymbol{\gamma}$ provided by Algorithm \ref{alg:cap}.

\vspace{-2mm}
\section{Experimental Results}
\vspace{-2mm}
\label{sec:experiment}
We compared our method with the works presented in \cite{mandelli2019facing} (denoted as M2019) and \cite{mandelli2020modified} (denoted as MFM), whose codes are available on git-hub 
We measure the method performance in terms of computational cost, True Positive Rates (TPR) for a fixed False Positive Rate $\text{FPR} \approx0.05$ and Area Under the Curve (AUC).
\par The dataset used for the experiments is VISION \cite{shullani2017vision}, from which we selected horizontal videos taken using the EIS (except videos from device D23 which have very low resolution $640\times 480$). Metadata were checked for every video to know whether it was taken upside down and, in this case, we rotated it by 180 degrees. 
For each device, we composed the image-based camera fingerprint with $L=100$ flat images. Similarly to \cite{mandelli2019facing}, it gets down-sampled and cropped, so to obtain a new fingerprint $\hat{\bt K}$  directly comparable to the test frames.
We estimated offline such down-scaling and crop parameters and report them in Table \ref{tab:scale_and_crop} for all tested devices.

We analysed between eight and twelve videos per device, for a total of 131 videos. Results were obtained on a server with the following characteristics: RAM 64GB, Processor Intel(R) Xeon(R) CPU E5-2630 v3 2.40GHz, GPU NVIDIA Tesla K40c 12 GB.

\vspace{-2mm}
\begin{table}[H]
\centering
\resizebox{\columnwidth}{!}{%
\begin{tabular}{c|cccccccccccc}
           & {\it D02}   & {\it D05}   & {\it D06}   & {\it D10}   & {\it D14}   & {\it D15}   & {\it D18}   & {\it D19}   & {\it D20}   & {\it D25}   & {\it D29}   & {\it D34}   \\ \toprule
$\lambda$ & 1.333 & 1.455 & 1.417 & 1.333 & 1.455 & 1.416 & 1.454 & 1.417 & 1.227 & 1.933 & 1.455 & 1.455 \\ \hline
$w_{tl}$   & 206   & 103   & 134   & 206   & 103   & 134   & 104   & 133   & 38    & 182   & 103   & 103   \\ \hline
$h_{tl}$   & 345   & 269   & 291   & 345   & 269   & 291   & 269   & 291   & 216   & 327   & 269   & 269   \\ \hline
$w_{br}$ & 2242  & 2140  & 2170  & 2242  & 2140  & 2170  & 2140  & 2170  & 2074  & 2218  & 2140  & 2140  \\ \hline
$h_{br}$   & 1491  & 1414  & 1437  & 1491  & 1414  & 1437  & 1414  & 1436  & 1362  & 1437  & 1414  & 1414  
\end{tabular}
}
\vspace{-2mm}
\caption{\footnotesize Estimated parameters for obtaining $\hat{\bt K}$ for each device starting from the image-based fingerprint.  $\lambda$ is the down-scaling parameters, and $(w_{\text{tl}},h_{\text{tl}})$ and $(w_{\text{br}},h_{\text{br}})$ are the coordinates of the top-left and bottom-right  corners of the rectangular crop.}
\label{tab:scale_and_crop}
\end{table}
\vspace{-3mm}
\begin{figure}[htpb!]
    \centering
\includegraphics[scale=0.37]{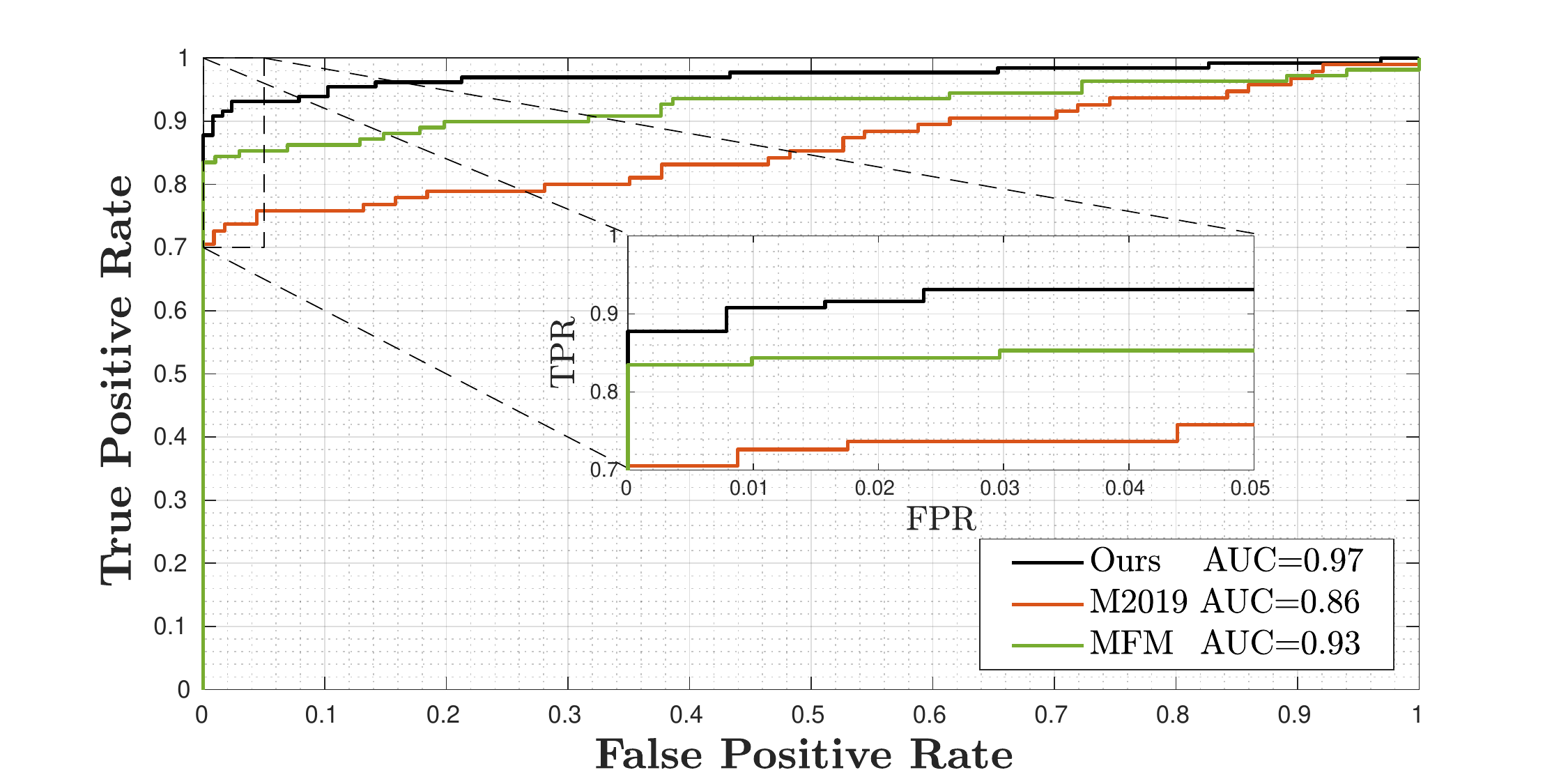}
\vspace{-2mm}
    \caption{\footnotesize ROC of the proposed method ``Ours", M2019 \cite{mandelli2019facing} and MFM \cite{mandelli2020modified}.}
    \label{fig:ROCs}
    \vspace{-5mm}
\end{figure}
\par In Figure \ref{fig:ROCs} we report ROC curves for our solution and state-of-the-art methods. We computed the PCE values used for Figure \ref{fig:ROCs} by matching all the 131 video with its device for H1, and with a different one for H0. The improvement  with respect to M2019 \cite{mandelli2019facing} and MFM \cite{mandelli2020modified} is evident both in terms of AUC and TPR, demonstrating the effectiveness of the proposed algorithm. 

A more detailed comparison at acquisition device level is shown in Table \ref{tab:TRP_time}, where we report the TPR for a fixed FPR=0.05 and the computational time required by each solution. These results prove the strong reduction of computational cost achieved with our strategy, {which shows a much smaller Elaboration Time Per Second (ETPS) of video, even if analysing up to three times more frames than \cite{mandelli2019facing, mandelli2020modified}}. {Conversely, the much higher ETPS of our method on the CPU proves the suitability of the GPU in problems with a large number of parameter combinations.} While we did not have the possibility to fully match the hardware described in \cite{mandelli2019facing} (which might allow for a faster application of their method), the time measures in Table \ref{tab:TRP_time} demonstrate the significantly lower computational and hardware constraints of our algorithm. 
The proposed method outperforms the state-of-the-art also in terms of TPR with FPR fixed to $0.05$, except for video sequences coming from the devices D06 and D25. We conjecture that their worse performance in terms of TPR is related to a noisy estimation of Eqs. \eqref{eq:momentum} and \eqref{eq:anchor_frames}, that we often observed on highly textured videos. However, we believe that with a more accurate keypoints selection, the identification results can be further improved and it is our purpose to investigate solutions similar to the ones proposed in \cite{battiato2007sift}.
\vspace{-2mm}
\section{Discussion and conclusions}
\label{sec:conclusions}

In this paper, we presented an innovative method for the identification of the source of EIS videos, where the more promising frames for this purpose are temporally localized. We did it by taking into account the inevitable temporal correlation of the EIS applied to neighbour frames and by defining a measure for the camera momentum. The results obtained so far on stabilized videos coming from VISION \cite{shullani2017vision} outperform previous approaches both in terms of identification accuracy and of computational efficiency.

However, we believe there is space for improvement in different aspects. 
In particular, we aim at improving the model used for the estimation of the camera momentum (e.g., by exploiting strategies described in \cite{battiato2007sift}) and to further optimize the use of the GPU for the EIS inversion in video frames. 
Another future direction could be to investigate strategies to incorporate deep networks (such as the Noiseprint \cite{cozzolino2019extracting} fingerprint extractor) in the forensics analysis, which have a high potential in improving algorithmic efficiency in testing but need to be adapted to deal with stabilization issues.
Moreover, moving from the hybrid scenario where the camera fingerprint is estimated on flat images to a fully video-based one were (potentially stabilized) videos are used for getting the reference fingerprints would be of high practical relevance.

\bibliographystyle{IEEEbib}
\bibliography{refs}

\begin{thebibliography}{10}

\bibitem{lukas2006digital}
J.~Lukas, J.~Fridrich, and M.~Goljan,
\newblock ``Digital camera identification from sensor pattern noise,''
\newblock {\em IEEE TIFS}, vol. 1, no. 2, pp. 205--214, 2006.

\bibitem{cozzolino2018noiseprint}
D.~Cozzolino and L.~Verdoliva,
\newblock ``Noiseprint: A cnn-based camera model fingerprint,''
\newblock {\em IEEE TIFS}, vol. 15, 2020.

\bibitem{dirik2007source}
A.~E. Dirik, H.~T. Sencar, and N.~Memon,
\newblock ``Source camera identification based on sensor dust
  characteristics,''
\newblock in {\em 2007 IEEE Workshop on Signal Processing Applications for
  Public Security and Forensics}. IEEE, 2007, pp. 1--6.

\bibitem{fridrich2009digital}
J.~Fridrich,
\newblock ``Digital image forensics,''
\newblock {\em IEEE Signal Processing Magazine}, vol. 26, no. 2, pp. 26--37,
  2009.

\bibitem{Goljian2012Sensor}
M.~Goljan and J.~Fridrich,
\newblock ``Sensor-fingerprint based identification of images corrected for
  lens distortion,''
\newblock in {\em Media Watermarking, Security, and Forensics 2012}.
  International Society for Optics and Photonics, 2012, vol. 8303, p. 83030H.

\bibitem{Goljan2008Digital}
M.~Goljan,
\newblock ``Digital camera identification from images--estimating false
  acceptance probability,''
\newblock in {\em International workshop on digital watermarking}. Springer,
  2008, pp. 454--468.

\bibitem{Darvish2019Camera}
M.~Darvish Morshedi~Hosseini and M.~Goljan,
\newblock ``Camera identification from hdr images,''
\newblock in {\em Proceedings of the ACM Workshop on Information Hiding and
  Multimedia Security}, 2019.

\bibitem{morimoto1998evaluation}
C.~Morimoto and R.~Chellappa,
\newblock ``Evaluation of image stabilization algorithms,''
\newblock in {\em Proceedings of the 1998 IEEE International Conference on
  Acoustics, Speech and Signal Processing, ICASSP'98 (Cat. No. 98CH36181)}.
  IEEE, 1998, vol.~5.

\bibitem{mandelli2019facing}
S.~Mandelli, P.~Bestagini, L.~Verdoliva, and S.~Tubaro,
\newblock ``Facing device attribution problem for stabilized video sequences,''
\newblock {\em IEEE TIFS}, vol. 15, pp. 14--27, 2019,
\newblock Available at
  \url{https://github.com/polimi-ispl/TIFS2019-stabilized-video-attribution}.

\bibitem{mandelli2020modified}
S.~Mandelli, F.~Argenti, P.~Bestagini, M.~Iuliani, A.~Piva, and S.~Tubaro,
\newblock ``A modified fourier-mellin approach for source device identification
  on stabilized videos,''
\newblock in {\em 2020 IEEE International Conference on Image Processing
  (ICIP)}. IEEE, 2020,
\newblock Available at
  \url{https://github.com/polimi-ispl/mfm-sdi-stabilized-videos}.

\bibitem{iuliani2019hybrid}
M.~Iuliani, M.~Fontani, D.~Shullani, and A.~Piva,
\newblock ``Hybrid reference-based video source identification,''
\newblock {\em Sensors}, vol. 19, no. 3, pp. 649, 2019.

\bibitem{cozzolino2019extracting}
D.~Cozzolino, G.~Poggi, and L.~Verdoliva,
\newblock ``Extracting camera-based fingerprints for video forensics,''
\newblock in {\em Proceedings of the IEEE Conference on Computer Vision and
  Pattern Recognition Workshops}, 2019, pp. 130--137.

\bibitem{bellavia2019prnu}
Fabio Bellavia, Massimo Iuliani, Marco Fanfani, Carlo Colombo, and Alessandro
  Piva,
\newblock ``Prnu pattern alignment for images and videos based on scene
  content,''
\newblock in {\em 2019 IEEE International Conference on Image Processing
  (ICIP)}. IEEE, 2019, pp. 91--95.

\bibitem{bellavia2021experiencing}
Fabio Bellavia, Marco Fanfani, Carlo Colombo, and Alessandro Piva,
\newblock ``Experiencing with electronic image stabilization and prnu through
  scene content image registration,''
\newblock {\em Pattern Recognition Letters}, vol. 145, pp. 8--15, 2021.

\bibitem{Chen2008Determining}
M.~Chen, J.~Fridrich, M.~Goljan, and J.~Luk{\'a}s,
\newblock ``Determining image origin and integrity using sensor noise,''
\newblock {\em IEEE TIFS}, vol. 3, no. 1, pp. 74--90, 2008.

\bibitem{Goljian2014Estimation}
M.~Goljan and J.~Fridrich,
\newblock ``Estimation of lens distortion correction from single images,''
\newblock in {\em Media Watermarking, Security, and Forensics 2014}.
  International Society for Optics and Photonics, 2014, vol. 9028, p. 90280N.

\bibitem{Goljan2009Large}
M.~Goljan, J.~Fridrich, and T.~Filler,
\newblock ``Large scale test of sensor fingerprint camera identification,''
\newblock {\em Proceedings of SPIE - The International Society for Optical
  Engineering}, February 2009.

\bibitem{cortiana2011performance}
A.~Cortiana, V.~Conotter, and F.~GB Boato, G.and De~Natale,
\newblock ``Performance comparison of denoising filters for source camera
  identification,''
\newblock in {\em Media Watermarking, Security, and Forensics III}.
  International Society for Optics and Photonics, 2011, vol. 7880, p. 788007.

\bibitem{amerini2009analysis}
I.~Amerini, R.~Caldelli, V.~Cappellini, F.~Picchioni, and A.~Piva,
\newblock ``Analysis of denoising filters for photo response non uniformity
  noise extraction in source camera identification,''
\newblock in {\em 2009 16th International Conference on Digital Signal
  Processing}. IEEE, 2009, pp. 1--7.

\bibitem{Mihcak1999Denoiser}
M.~K. Mihcak, I.~Kozintsev, K.~Ramchandran, and P.~Moulin,
\newblock ``Low-complexity image denoising based on statistical modeling of
  wavelet coefficients,''
\newblock {\em IEEE Signal Processing Letters (SPL)}, vol. 6, pp. 300--303,
  1999.

\bibitem{Kang12}
X.~Kang, Y.~Li, Z.~Qu, and J.~Huang,
\newblock ``Enhancing source camera identification performance with a camera
  reference phase sensor pattern noise,''
\newblock {\em IEEE TIFS}, vol. 7, no. 2, pp. 393--402, 2012.

\bibitem{Goljan2008Camera}
M.~Goljan and J.~Fridrich,
\newblock ``Camera identification from cropped and scaled images,''
\newblock in {\em Security, Forensics, Steganography, and Watermarking of
  Multimedia Contents X}. International Society for Optics and Photonics, 2008,
  vol. 6819, p. 68190E.

\bibitem{lowe2004distinctive}
David~G. Lowe,
\newblock ``Distinctive image features from scale-invariant keypoints,''
\newblock {\em International journal of computer vision}, vol. 60, no. 2, pp.
  91--110, 2004.

\bibitem{chum2005two}
O.~Chum, T.~Werner, and J.~Matas,
\newblock ``Two-view geometry estimation unaffected by a dominant plane,''
\newblock in {\em 2005 IEEE Computer Society Conference on Computer Vision and
  Pattern Recognition (CVPR'05)}. IEEE, 2005, vol.~1.

\bibitem{battiato2007sift}
S.~Battiato, G.~Gallo, G.~Puglisi, and S.~Scellato,
\newblock ``Sift features tracking for video stabilization,''
\newblock in {\em 14th international conference on image analysis and
  processing (ICIAP 2007)}. IEEE, 2007.

\bibitem{bundy1984breadth}
A.~Bundy and L.~Wallen,
\newblock ``Breadth-first search,''
\newblock in {\em Catalogue of artificial intelligence tools}, pp. 13--13.
  Springer, 1984.

\bibitem{shullani2017vision}
D.~Shullani, M.~Fontani, M.~Iuliani, O.~Al~Shaya, and A.~Piva,
\newblock ``Vision: a video and image dataset for source identification,''
\newblock {\em EURASIP Journal on Information Security}, vol. 2017, no. 1, pp.
  1--16, 2017.

\end{thebibliography}

\end{document}